\documentclass[10pt,twocolumn,letterpaper]{article}

\usepackage[pagenumbers]{cvpr} 
\usepackage{graphicx}
\usepackage{multirow}
\usepackage{colortbl}
\usepackage{microtype}
\definecolor{cvprblue}{rgb}{0.21,0.49,0.74}
\usepackage[pagebackref,breaklinks,colorlinks,allcolors=cvprblue]{hyperref}

\def\confName{CVPR}
\def\confYear{2027}

\title{Real-Anchored Complementary Evidence for Generalizable \\AI-Generated Image Detection%
\iftoggle{cvprfinal}{%
    \thanks{This work was supported in part by the National Natural Science
    Foundation of China under Grant 62502308, in part by the Shanghai PuJiang
    Project under Grant 24PJA056, in part by the Natural Science Foundation
    of Shanghai under Grant 25ZR1402266, in part by the Open Research Fund
    of the State Key Laboratory of Blockchain and Data Security, Zhejiang
    University, under Grant A2536, and in part by Phase IV of the
    Cybersecurity Student Innovation Funding Program.}%
}{}%
}

\author{%
    Wenpeng~Mu \quad
    Tanfeng~Sun \quad
    Zhenyu~Shao \quad
    Qiang~Xu\thanks{Corresponding author.} \quad
    Xinghao~Jiang\\
    School of Computer Science, Shanghai Jiao Tong University\\
    Shanghai, China\\
    {\small\ttfamily
     \{wpmu,tfsun,young.s,xuqiangwhu,xhjiang\}@sjtu.edu.cn}%
}

\begin{document}
\maketitle
\begin{abstract}
Evolving generative models challenge detector generalization, as methods trained on generated examples may rely on generator-specific cues. By contrast, authentic photographs share regularities rooted in physical image formation and acquisition, providing a common reference beyond individual generators. Building on these regularities, we propose Real-Anchored Complementary Evidence (RACE), which learns complementary global and local detection evidence using only authentic training images. First, the global pathway combines frozen and learned representations to assess proximity to authentic references. Next, it incorporates responses to controlled perturbations, complementing static compatibility with representation sensitivity. The local pathway then reconstructs complete trace descriptors, using intermediate features and signed perturbation responses as context while restricting detection evidence to static terminal-residual reconstruction errors. Finally, authentic-image statistics align the global and local scores for fusion, and an independent authentic set calibrates a decision threshold held fixed across test generators. Across six benchmarks, RACE achieves the highest average AUROC of 96.59\% among the compared methods, reaching 97.18\% on five out-of-domain benchmarks. Together with robustness to image perturbations and variation in authentic-image sources, these results support authentic-image regularities as a transferable basis for detection across diverse generators and image conditions.
\end{abstract}
\section{Introduction}

Recent advances in generative artificial intelligence (GAI) have made photorealistic image synthesis increasingly accessible, complicating authenticity verification through. Although these capabilities support numerous legitimate applications, they can also facilitate disinformation, fraud, and identity impersonation, threatening information integrity and trust in digital media~\cite{xu2025advancements}. Reliable detection of AI-generated images is therefore essential. Meeting this need in practice requires detectors to generalize beyond the synthesis methods encountered during development and remain effective as generators and deployment conditions evolve.

\begin{figure}[t]
    \centering
    \includegraphics[width=\columnwidth]{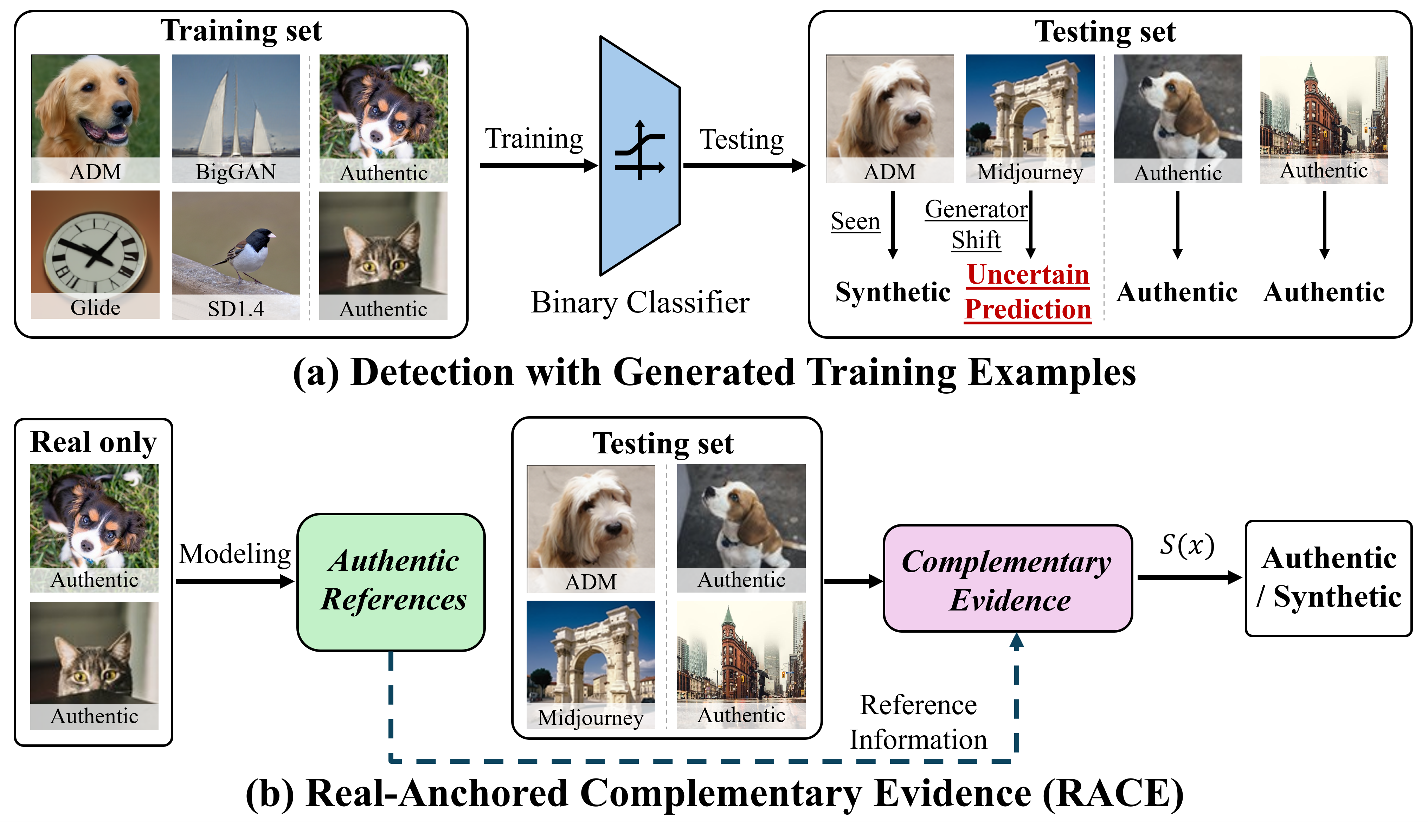}
    \caption{Comparison of detection paradigms.
    (a) Detectors trained with generated examples may become unreliable under generator shifts.
    (b) RACE assesses authenticity through complementary global and local evidence anchored in authentic images.}
    \label{fig:first}
\end{figure}

Many existing detectors formulate detection as supervised binary classification using authentic and generated training images, as illustrated in Fig.~\ref{fig:first}(a). Representative approaches employ data augmentation~\cite{cvpr/WangW0OE20}, reconstruction errors~\cite{iccv/WangBZWHCL23}, neighboring-pixel relationships~\cite{cvpr/TanLZWGLW24}, and pretrained or hybrid representations~\cite{cvpr/OjhaLL23,iclr/YanLCHJ0X25}. However, their decision boundaries remain tied to the synthetic distributions observed during training, making them susceptible to overfitting to generator-specific artifacts. As generators evolve, these artifacts can change or disappear, undermining detection reliability even for models that perform well on familiar generators.

In contrast, modeling authentic-image regularities has emerged as a promising direction for generalizable detection~\cite{corr/abs-2602-02222,corr/abs-2606-06918,corr/abs-2601-18900}. This shifts the emphasis from learning generator-specific discriminative patterns to assessing conformity with authentic-image distributions. Existing approaches typically model selected aspects of these distributions. However, conformity in the modeled aspects does not guarantee consistency in others, motivating an assessment that draws on complementary global and local evidence. This raises our question: \textbf{How can we learn such complementary evidence from authentic images alone?}

To address this question, we propose Real-Anchored Complementary Evidence (RACE), a framework that integrates complementary global and local authenticity evidence, as illustrated in Fig.~\ref{fig:first}(b). Its added modules are trained exclusively on authentic images. First, the global pathway uses Complementary Feature Encoding (CFE) to pair frozen DINOv2~\cite{oquab2023dinov2} features with learned CLIP~\cite{radford2021learning} representations, assessing proximity to authentic references. Next, since static proximity does not establish consistent behavior under input changes, Perturbation Response Profiling (PRP) measures representation sensitivity through controlled perturbations. The local pathway then complements these global assessments with Noiseprint Reconstruction with Selective Readout (NP-RSR), examining local trace discrepancies that global representations can overlook. It reconstructs complete trace descriptors but scores only static terminal-residual errors, using the remaining information as reconstruction context. Finally, authentic-image statistics align the global and local scores for fusion, and an independent authentic set calibrates a decision threshold held fixed across test generators.

Extensive experiments on six challenging image benchmarks evaluate RACE across GANs, diffusion models, and visual autoregressive models, as well as recent high-performance closed-source generators such as GPT Image 2 and Nano Banana 2. Compared with 13 baselines, RACE achieves the highest average AUROC of 96.59\% across all six benchmarks. On the five out-of-domain benchmarks, it ranks first in both average AUROC and AP, achieving 97.18\% and 96.17\%, respectively, demonstrating strong transfer across diverse generation paradigms. Further evaluations show robust performance under image perturbations and across different authentic-image sources, while ablation studies substantiate the complementary contributions of global and local evidence.

We summarize our main contributions as follows:
\begin{itemize}
    \item \textbf{Real-Anchored Global and Local Evidence.}
    We propose RACE, a framework that unifies authentic-reference proximity, controlled perturbation responses, and local trace reconstruction for complementary authenticity assessment. Its added modules are trained exclusively on authentic images, which also provide the statistical reference for evidence fusion and threshold calibration, establishing a detection rule held fixed across test generators.

    \item \textbf{Perturbation Response Profiling.}
    We develop PRP to characterize the sensitivity of frozen representations through Gaussian, uniform, and structured high-frequency probes. Normalizing these responses against authentic-image statistics yields a sensitivity profile that complements static reference proximity, extending global assessment from feature compatibility to perturbation behavior without additional trainable parameters.

    \item \textbf{Local Reconstruction with Selective Readout.}
    We develop NP-RSR to decouple descriptor reconstruction from anomaly scoring. It reconstructs complete local trace descriptors, with intermediate activations and signed perturbation responses providing context, while restricting detection to static terminal-residual reconstruction errors. This design retains rich information for reconstruction while concentrating local detection evidence on residual discrepancies.
\end{itemize}

\section{Related Work}

\subsection{Discriminative Methods for AI-Generated \\Image Detection}

AI-generated image detection has progressed from statistical cues, such as saturation~\cite{icip/McCloskeyA19} and pixel co-occurrence~\cite{NatarajMMCFBR19}, to learned forensic representations. CNNSpot~\cite{cvpr/WangW0OE20} emphasizes data augmentation, while DIRE~\cite{iccv/WangBZWHCL23} and NPR~\cite{cvpr/TanLZWGLW24} exploit reconstruction errors and neighboring-pixel relationships, respectively. UnivFD~\cite{cvpr/OjhaLL23} uses pretrained features for classification, and MF$^2$DA~\cite{mu2026mf2da} combines CLIP semantics with local edge artifacts. DRCT~\cite{icml/ChenZYY24} and DDA~\cite{nips/ChenXYZWXCXGYD25} improve training data through diffusion reconstruction and joint pixel--frequency alignment, respectively. Veritas~\cite{iclr/abs-2508-21048} further introduces pattern-aware reasoning for interpretable detection. Despite these advances, their discriminative criteria remain influenced by the generated examples available during development.

\subsection{Authentic-Image Modeling for AI-Generated Image Detection}

Authentic-image modeling remains comparatively less explored than supervised real--synthetic classification. Representative studies include MIRROR~\cite{corr/abs-2602-02222}, which reconstructs reference features from authentic prototypes but still requires generated-sample supervision for its detection stage. RIGID~\cite{corr/abs-2405-20112} and ConV~\cite{zhang2026detecting} assess transformation consistency in pretrained representations, while DRIFT~\cite{corr/abs-2606-06918} learns robust and fragile subspaces tailored to different transformation families. But they do not explicitly assess static proximity to empirical authentic-reference neighborhoods. RealStats~\cite{corr/abs-2601-18900} calibrates and aggregates multiple statistics using authentic distributions; however, this statistical framework relies on the evidence supplied by its underlying detectors rather than learning additional forensic representations. RACE integrates authentic-reference proximity, perturbation responses, and local trace evidence through authentic-only learning and calibration.

\section{Method}
\label{sec:race-method}

\begin{figure*}
    \centering
    \includegraphics[width=\linewidth]{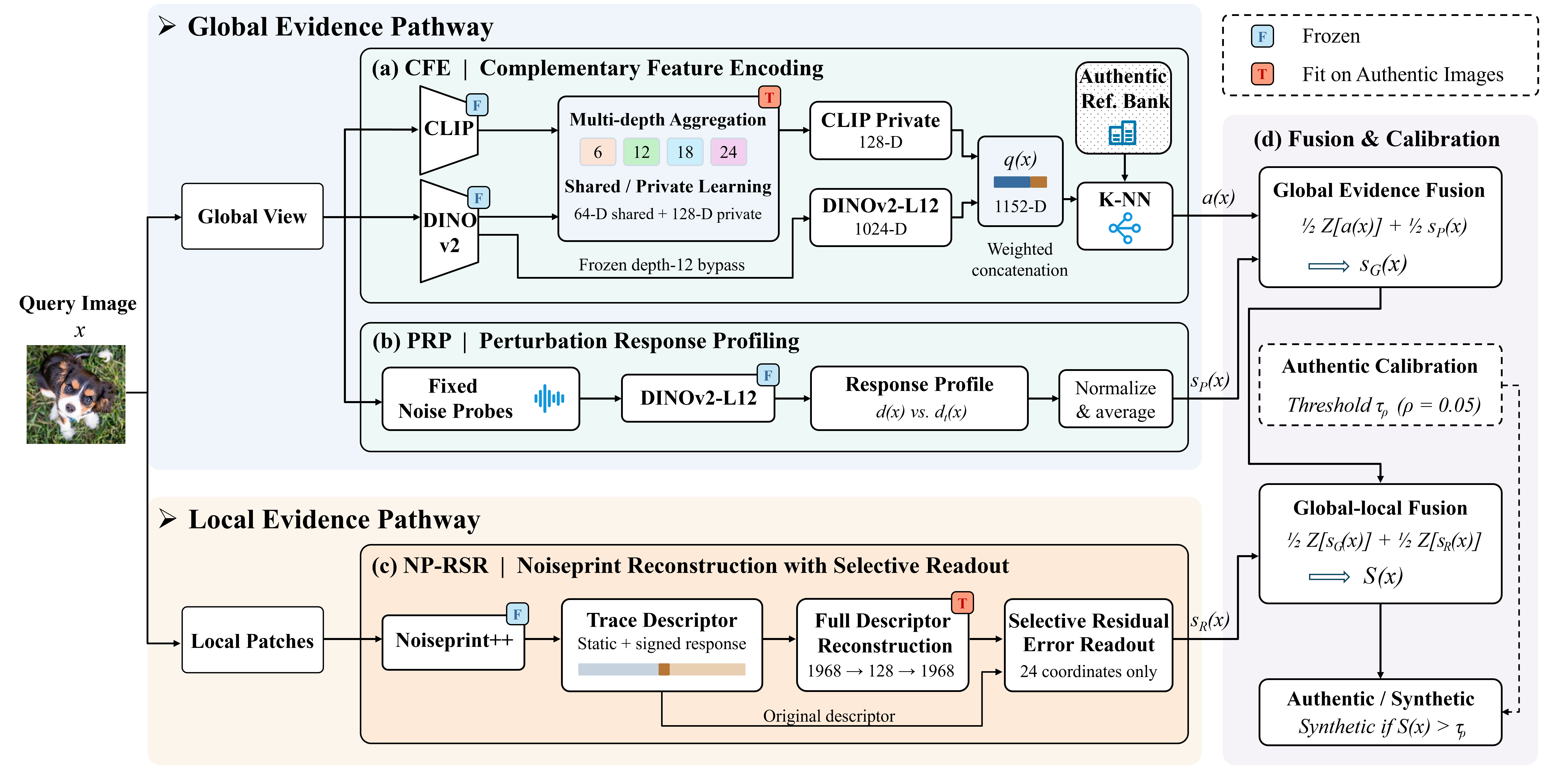}
    \caption{Overview of RACE. The global pathway combines authentic-reference proximity (CFE) and perturbation responses (PRP), while the local pathway reconstructs trace descriptors with selective residual scoring (NP-RSR). The scores are fused, and the decision threshold is calibrated using authentic images.}
    \label{fig:framework}
\end{figure*}

\subsection{Motivation}
Human vision interprets new observations using regularities acquired from experience~\cite{series2013learning,rao1999predictive}. Natural-image datasets also exhibit low intrinsic dimensionality relative to their pixel spaces~\cite{iclr/PopeZAGG21}, while neighborhood relationships can reveal underlying geometric structure under suitable sampling assumptions~\cite{tenenbaum2000global}. These observations suggest using regularities captured by authentic-image representations as an empirical reference for authenticity assessment. However, agreement with authentic references in a global feature space does not by itself establish consistency in perturbation behavior or local residual statistics. We therefore hypothesize that \textbf{global reference proximity, perturbation responses, and local trace consistency provide complementary evidence of departures from authentic-image regularities}. Jointly assessing these aspects can reveal discrepancies overlooked by an individual assessment, supporting detection without explicitly modeling individual generators.

\subsection{Overview of RACE}
\label{sec:race-framework}
As shown in Figure~\ref{fig:framework}, Real-Anchored Complementary Evidence (RACE) integrates a global pathway comprising Complementary Feature Encoding (CFE) for authentic-reference retrieval (Section~\ref{sec:race-cfe}) and Perturbation Response Profiling (PRP) for perturbation-response assessment (Section~\ref{sec:race-prp}), with a local pathway using Noiseprint Reconstruction with Selective Readout (NP-RSR) for trace reconstruction and selective residual scoring (Section~\ref{sec:race-nprsr}). The resulting scores are normalized and fused, followed by threshold calibration on an independent authentic set (Section~\ref{sec:race-fusion}).

\subsection{Complementary Feature Encoding}
\label{sec:race-cfe}
Authentic-reference detection requires representations that preserve informative variation within authentic images. CLIP~\cite{radford2021learning} aligns images with natural-language descriptions, emphasizing semantic concepts and attributes associated with language. DINOv2~\cite{oquab2023dinov2} learns from images through image- and patch-level self-supervision, capturing visual appearance and structure without textual alignment. These different training objectives can induce distinct feature neighborhoods, allowing visual distinctions underrepresented by one encoder to remain informative in the other. We therefore combine their representations for authentic-reference retrieval. With both backbones frozen, CFE pairs a DINOv2 reference anchor with a representation learned from CLIP features, preserving pretrained visual geometry while incorporating complementary variation learned from authentic images.

\paragraph{Adaptive multi-depth aggregation.}
For encoder $b\in\{C,D\}$, corresponding to CLIP and DINOv2, respectively, let $\mathbf{f}_{\ell}^{b}(x)$ denote the CLS feature extracted from the global view $g(x)$ at depth $\ell\in\mathcal{L}=\{6,12,18,24\}$. Each encoder uses a transformation $e_b$ shared across depths and separate gates for the shared and private routes, indexed by $r\in\{\mathrm{sh},\mathrm{pr}\}$. Each gate takes the mean of the layer-normalized depth features as input and produces softmax aggregation weights $\boldsymbol{\pi}_{r}^{b}(x)$. The aggregated features are incorporated into the corresponding frozen anchors through bounded residual adaptation:
\begin{equation}
\begin{aligned}
\mathbf{v}_{r}^{b}(x)
&=
\sum_{\ell\in\mathcal{L}}
\pi_{r,\ell}^{b}(x)e_b(\mathbf{f}_{\ell}^{b}(x)),\\
\widetilde{\mathbf{f}}_{r}^{b}(x)
&=
\mathbf{f}_{\ell_r}^{b}(x)
+
\eta\tanh(\gamma_{r}^{b})\,
\operatorname{sg}\!\left(
\|\mathbf{f}_{\ell_r}^{b}(x)\|_2
\right)
\nu(\mathbf{v}_{r}^{b}(x)),
\end{aligned}
\end{equation}
where $\nu$ denotes $\ell_2$ normalization, $\operatorname{sg}$ stops gradients, $\ell_{\mathrm{sh}}=24$, $\ell_{\mathrm{pr}}=12$, and $\eta=0.25$. Initializing $\gamma_{r}^{b}=0$ recovers the frozen anchor at the start of training. Throughout training, the residual magnitude is bounded by $\eta\|\mathbf{f}_{\ell_r}^{b}(x)\|_2$.

\paragraph{Shared and private representation learning.}
We learn shared and private codes to capture cross-encoder agreement while retaining encoder-specific variation. After coordinate-wise standardization using authentic training statistics, projection heads map the adapted features to a 64-dimensional shared code $\mathbf{c}_{\mathrm{sh}}^{b}$ and a 128-dimensional private code $\mathbf{c}_{\mathrm{pr}}^{b}$. Using two augmented views of each authentic image, we optimize
\begin{equation}
\begin{aligned}
\mathcal{L}_{\mathrm{CFE}}
={}&
10\mathcal{L}_{\mathrm{view,sh}}
+
2.5\mathcal{L}_{\mathrm{view,pr}}
+
10\mathcal{L}_{\mathrm{align}}\\
&+
25\mathcal{L}_{\mathrm{var},C}
+
\mathcal{L}_{\mathrm{cov},C}
+
\mathcal{L}_{\mathrm{sep}}\\
&+
5\mathcal{L}_{\mathrm{rec},C}
+
\mathcal{L}_{\mathrm{geo},C}.
\end{aligned}
\end{equation}
The view-consistency terms encourage agreement between augmented views in both routes, whereas cross-encoder alignment is applied only to the shared codes. The CLIP-side variance and covariance penalties encourage code diversity and reduce within-code redundancy~\cite{bardes2021vicreg}, while $\mathcal{L}_{\mathrm{sep}}$ discourages redundancy between shared and private codes. The CLIP-side reconstruction and geometry terms further encourage retention of frozen-anchor information and pairwise cosine relationships, respectively.

\paragraph{Complementary reference proximity.}
For retrieval, we combine a frozen reference anchor with a complementary representation learned from authentic images. Specifically, we pair the unit-normalized frozen DINOv2 depth-12 feature $\mathbf{d}(x)$ with the normalized CLIP private code $\mathbf{c}(x)=\nu(\mathbf{c}_{\mathrm{pr}}^{C}(x))$:
\begin{equation}
\mathbf{q}(x)
=
\nu\!\left(
\left[
\sqrt{\alpha_D}\mathbf{d}(x);
\sqrt{\alpha_C}\mathbf{c}(x)
\right]
\right)
\in\mathbb{R}^{1152}.
\end{equation}
The remaining learned codes serve only as training auxiliaries.

To account for differences in the dispersion of authentic-neighbor distances, we estimate the component weights on the authentic development set $\mathcal{D}_{\mathrm{dev}}$ as $\alpha_b=\omega_b^{-1}/(\omega_D^{-1}+\omega_C^{-1})$, where $\omega_b$ is the IQR-based scale of mean $K$-neighbor distances in the corresponding unit-normalized authentic-reference space. These weights remain fixed during calibration and inference.

Given the reference bank $\mathcal{B}=\{\mathbf{q}(x_i):x_i\in\mathcal{D}_{\mathrm{ref}}\}$, let $\mathcal{N}_K(\mathbf{q}(x);\mathcal{B})$ denote the indices of its $K$ nearest neighbors. We define the CFE score as the mean cosine distance to these neighbors:
\begin{equation}
a(x)
=
\frac{1}{K}
\sum_{i\in\mathcal{N}_K(\mathbf{q}(x);\mathcal{B})}
\left[
1-\mathbf{q}(x)^{\top}\mathbf{q}(x_i)
\right],
\qquad K=16.
\end{equation}
For unit-normalized representations, Euclidean and cosine distances yield identical neighbor rankings. A larger $a(x)$ indicates weaker compatibility with the available authentic references in the complementary representation space.

\subsection{Perturbation Response Profiling}
\label{sec:race-prp}
Static feature proximity provides only one view of the variation within authentic images. To enrich this characterization, PRP probes how frozen representations respond to controlled input perturbations, revealing differences in sensitivity that static embeddings alone may not capture. These responses are assessed against authentic-image statistics, providing complementary evidence beyond reference proximity without additional trainable parameters.

\paragraph{Controlled noise perturbations.}
Specifically, the global $224\times224$ RGB view $g(x)$ is perturbed using $\mathcal{T}=\{\mathrm{Gau},\mathrm{Uni},\mathrm{HH}\}$, comprising zero-mean Gaussian, equal-variance uniform, and Haar HH noise. The perturbed view is
$T_t(g(x))=\operatorname{clip}_{[0,1]}(g(x)+\epsilon_t)$,
with all probes having the same nominal RGB standard deviation of $\sigma=1/255$ before clipping. The Haar HH probe combines Gaussian coefficients with a $2\times2$ checkerboard pattern to examine responses to structured high-frequency perturbations. One noise field is sampled for each probe type and reused across development, calibration, and inference. All perturbations are applied in RGB space before DINOv2 normalization.

\paragraph{Representation response and profiling.}
We compute perturbation responses using the frozen DINOv2 depth-12 representation that also serves as the reference anchor in CFE. Let $\mathbf{d}(x)$ and $\mathbf{d}_t(x)$ denote the unit-normalized features of the original and perturbed views, respectively. Their cosine distance measures the representation response:
\begin{equation}
    r_t(x)
    =
    1-\mathbf{d}_t(x)^\top\mathbf{d}(x).
    \label{eq:race-probe-response}
\end{equation}
To account for differences in the authentic response distributions across probes, we normalize each response using its own development statistics. Specifically, we define
$\mathcal Z_{r_t}(r)=(r-m_{r_t})/b_{r_t}$,
where $m_{r_t}$ and $b_{r_t}$ are the median and numerically floored IQR-based scale estimated on the authentic development set $\mathcal D_{\mathrm{dev}}$. The normalized responses are then averaged to obtain the PRP score:
\begin{equation}
    s_{\mathrm P}(x)
    =
    \frac{1}{|\mathcal T|}
    \sum_{t\in\mathcal T}
    \mathcal Z_{r_t}(r_t(x)).
    \label{eq:race-response-profile}
\end{equation}
All normalization statistics remain fixed during calibration and inference. A larger $s_{\mathrm P}(x)$ indicates greater average normalized sensitivity relative to the authentic baseline.

\subsection{Noiseprint Reconstruction with Selective \\Readout}
\label{sec:race-nprsr}

NP-RSR complements global feature assessment with explicit modeling of local acquisition and processing traces. Using a frozen Noiseprint++ backbone~\cite{guillaro2023trufor}, it learns to reconstruct trace descriptors from authentic images while restricting anomaly scoring to static terminal-residual errors.

\paragraph{Local trace descriptor construction.}
We extract $M=3$ native $128\times128$ patches from the center, top-left, and bottom-right of each image. Images with a short side below 128 pixels are first resized to a short side of 256 pixels. For each patch $u$, the static descriptor
$\psi(u)=[\psi_{\mathrm{int}}(u);\psi_{\mathrm{res}}(u)]\in\mathbb{R}^{984}$
combines 960 intermediate-activation statistics and 24 terminal-residual statistics. Intermediate statistics summarize convolutions 6, 11, and 16 using $2\times2$ average pooling and spatial standard deviations. Residual statistics capture distributions, directional autocorrelations, phase-conditioned energies, and normalized radial spectral energies. All statistics exclude a 16-pixel boundary.

We append signed descriptor responses to a fixed local Gaussian probe $\delta$ with RGB standard deviation $1/255$, distinct from the global PRP probes:
\begin{equation}
\begin{aligned}
\Delta\psi(u)
&=
\psi\!\left(\operatorname{clip}_{[0,1]}(u+\delta)\right)
-\psi(u),\\
v_m(x)
&=
[\psi(u_m(x));\Delta\psi(u_m(x))]
\in\mathbb{R}^{1968}.
\end{aligned}
\label{eq:race-full-descriptor}
\end{equation}
The complete descriptor is standardized as
$z_m=\operatorname{clip}_{[-15,15]}((v_m-\mu_{\mathrm{tr}})/\sigma_{\mathrm{tr}})$,
using fixed coordinate means $\mu_{\mathrm{tr}}$ and numerically floored standard deviations $\sigma_{\mathrm{tr}}$ estimated from authentic training patches.

\paragraph{Masked descriptor reconstruction.}
We train an encoder $A_\phi$ and decoder $B_\phi$ on authentic descriptors to reconstruct complete inputs from masked observations. Independent coordinate masking gives
$\widetilde z_i=z_i\odot m_i/(1-p)$,
where $m_{ij}\sim\operatorname{Bernoulli}(1-p)$ and $p=0.2$.
The reconstruction $\widehat z_i=B_\phi(A_\phi(\widetilde z_i))$ is optimized against the full unmasked descriptor:
\begin{equation}
\begin{aligned}
\mathcal{L}_{\mathrm{NP}}
={}&
\frac{1}{BD}
\sum_{i=1}^{B}
\|\widehat z_i-z_i\|_2^2\\
&+
0.1\mathcal{L}_{\mathrm{geo,NP}}
+
0.1\mathcal{L}_{\mathrm{var,NP}}
+
0.001\mathcal{L}_{\mathrm{cov,NP}},
\end{aligned}
\label{eq:race-np-objective}
\end{equation}
where $B$ is the patch batch size and all $D=1968$ coordinates receive equal reconstruction weight. The regularizers encourage preservation of descriptor pairwise cosine geometry in the latent space, penalize latent standard deviations below 0.5, and reduce off-diagonal covariance, respectively.

\paragraph{Selective residual readout.}
At inference, we reconstruct each complete, unmasked descriptor as
$\widehat z_m=B_\phi(A_\phi(z_m))$.
Let $\Pi_{\mathrm R}\in\mathbb{R}^{24\times1968}$ select the 24 static terminal-residual coordinates. The image-level score averages their reconstruction errors across patches:
\begin{equation}
s_{\mathrm R}(x)
=
\frac{1}{24M}
\sum_{m=1}^{M}
\left\|
\Pi_{\mathrm R}
\bigl(\widehat z_m(x)-z_m(x)\bigr)
\right\|_2^2.
\label{eq:race-selective-readout}
\end{equation}
Intermediate activations and signed responses thus provide reconstruction context, while their reconstruction errors are excluded from the anomaly score.

\subsection{Evidence Fusion and Authentic Calibration}
\label{sec:race-fusion}

We normalize scores using $\mathcal Z_f(t)=(t-m_f)/b_f$, where $m_f$ and $b_f$ are the median and numerically floored IQR-based scale estimated on the authentic development set $\mathcal D_{\mathrm{dev}}$. The global and final scores are computed in two stages:
\begin{equation}
\begin{aligned}
s_{\mathrm G}(x)
&=
\frac{1}{2}
\left[
\mathcal Z_a(a(x))+s_{\mathrm P}(x)
\right],\\
S(x)
&=
(1-\lambda)\mathcal Z_{s_{\mathrm G}}(s_{\mathrm G}(x))
+
\lambda\mathcal Z_{s_{\mathrm R}}(s_{\mathrm R}(x)).
\end{aligned}
\label{eq:race-final-fusion}
\end{equation}
Where $\lambda=0.5$. All normalization statistics, component weights, and fusion coefficients are fixed before threshold calibration and inference.

\paragraph{Threshold calibration using authentic images.}
With $S$ fixed, we calibrate the decision threshold on $n$ independent authentic images. For a nominal false-positive budget $\rho$, let $j_\rho=\lceil(n+1)(1-\rho)\rceil$ and let $S_{(j)}^{\mathrm{cal}}$ denote the $j$th smallest calibration score. We set
\begin{equation}
\tau_\rho
=
\begin{cases}
S_{(j_\rho)}^{\mathrm{cal}},
& j_\rho\le n,\\
+\infty,
& j_\rho=n+1.
\end{cases}
\label{eq:race-calibrated-decision}
\end{equation}
The prediction is $\widehat y(x)=\mathbf 1\{S(x)>\tau_\rho\}$, with 1 denoting synthetic and 0 authentic. We use $n=5000$ and $\rho=0.05$.

The nominal false-positive interpretation requires exchangeability between calibration and future authentic observations~\cite{10.1561/2200000101}. Both the score and threshold remain fixed across test generators, and source-specific false-positive rates are evaluated separately.

\section{Experiments}

\subsection{Experimental Setup}

\begin{table*}[t]
\centering
\caption{Comparison on six datasets in terms of AUROC (\%) and AP (\%) ($\uparrow$).
Results within each dataset are averaged over its subsets; Average is the macro-average over all six datasets, including GenImage.
The best and second-best results in each metric column are shown in \textbf{bold} and \underline{underlined}, respectively.
The vertical rule separates GenImage from the out-of-domain datasets.
$\dagger$ indicates methods that are not trained on GenImage.}
\label{tab:race_generalization}
\begingroup
\setlength{\tabcolsep}{3pt}
\renewcommand{\arraystretch}{1.15}
\resizebox{\textwidth}{!}{%
\begin{tabular}{@{}ll|cc|*{5}{cc}|cc@{}}
\hline\hline
\multirow{2}{*}{Method} & \multirow{2}{*}{Venue}
& \multicolumn{2}{|c|}{GenImage}
& \multicolumn{2}{c}{ARForensics}
& \multicolumn{2}{c}{QwenImageBench}
& \multicolumn{2}{c}{EvalGEN}
& \multicolumn{2}{c}{ForenSynths}
& \multicolumn{2}{c}{AIGIBench}
& \multicolumn{2}{|c}{Average} \\
\cline{3-16}
& & AUROC & AP & AUROC & AP & AUROC & AP
& AUROC & AP & AUROC & AP & AUROC & AP & AUROC & AP \\
\hline
CNNSpot & CVPR 2020 & 85.10 & 84.69 & 71.93 & 71.32 & 46.43 & 50.99 & 87.29 & 89.16 & 56.65 & 53.84 & 65.19 & 62.19 & 68.77 & 68.70 \\
UnivFD & CVPR 2023 & 94.35 & 94.49 & 88.55 & 88.07 & 90.25 & 89.39 & 95.07 & 94.82 & 92.62 & 93.61 & 93.94 & 93.20 & 92.46 & 92.26 \\
DIRE & ICCV 2023 & 95.07 & 94.84 & 88.27 & 86.85 & 70.19 & 70.64 & 81.38 & 82.64 & 80.49 & 76.79 & 84.78 & 81.27 & 83.36 & 82.17 \\
NPR & CVPR 2024 & 93.91 & 94.09 & 86.14 & 86.21 & 58.08 & 62.91 & 68.06 & 70.99 & 76.64 & 77.50 & 72.58 & 72.75 & 75.90 & 77.41 \\
DRCT & ICML 2024 & \underline{99.59} & \underline{99.54} & 97.68 & 97.31 & 77.45 & 75.66 & 96.25 & 95.28 & 94.08 & 93.59 & 95.70 & 94.56 & 93.46 & 92.66 \\
D$^{3}$QE & ICCV 2025 & 91.71 & 91.30 & 86.15 & 84.73 & 85.16 & 85.65 & 92.90 & 92.33 & 86.83 & 87.50 & 80.96 & 80.90 & 87.28 & 87.07 \\
LOTA & ICCV 2025 & 93.02 & 93.29 & 91.39 & 91.67 & 56.22 & 62.72 & 34.62 & 42.87 & 89.36 & 90.06 & 80.40 & 82.80 & 74.17 & 77.24 \\
DDA & NeurIPS 2025 & \textbf{99.77} & \textbf{99.76} & \underline{99.64} & \underline{99.63} & 81.32 & 83.88 & 98.31 & 98.18 & \underline{97.71} & \textbf{98.33} & 96.42 & 96.42 & \underline{95.53} & \textbf{96.03} \\
MF$^{2}$DA & TIFS 2026 & 97.13 & 97.18 & 93.07 & 92.78 & 85.78 & 86.54 & 95.19 & 95.48 & 93.03 & 94.41 & 94.96 & 95.08 & 93.19 & 93.58 \\
IAPL & CVPR 2026 & 96.89 & 95.80 & \textbf{99.95} & \textbf{99.95} & 87.64 & 87.62 & 90.06 & 90.20 & 96.49 & 96.22 & \underline{98.67} & \textbf{98.36} & 94.95 & 94.69 \\
\hline 
ConV$^\dagger$ & NeurIPS 2025 & 76.32 & 76.58 & 65.29 & 64.74 & 71.63 & 70.18 & 79.56 & 78.45 & 71.58 & 71.27 & 70.31 & 67.20 & 72.45 & 71.40 \\
RealStats$^\dagger$ & AISTATS 2026 & 60.97 & 61.48 & 44.96 & 47.29 & 44.80 & 46.35 & 43.07 & 45.44 & 75.39 & 74.42 & 65.73 & 63.70 & 55.82 & 56.45 \\
Veritas$^\dagger$ & ICLR 2026 & 79.75 & 79.51 & 77.21 & 76.89 & \textbf{95.94} & \textbf{95.59} & \textbf{99.60} & \textbf{99.40} & 95.28 & 94.23 & 95.28 & 92.79 & 90.51 & 89.74 \\
\rowcolor[gray]{0.92}
RACE$^\dagger$ & / & 93.59 & 91.78 & 94.69 & 92.79 & \underline{95.34} & \underline{93.45} & \underline{99.09} & \underline{98.81} & \textbf{98.06} & \underline{97.56} & \textbf{98.74} & \underline{98.25} & \textbf{96.59} & \underline{95.44} \\
\hline\hline
\end{tabular}%
}
\endgroup
\end{table*}

\begin{figure}[t]
    \centering
    \includegraphics[width=\linewidth]{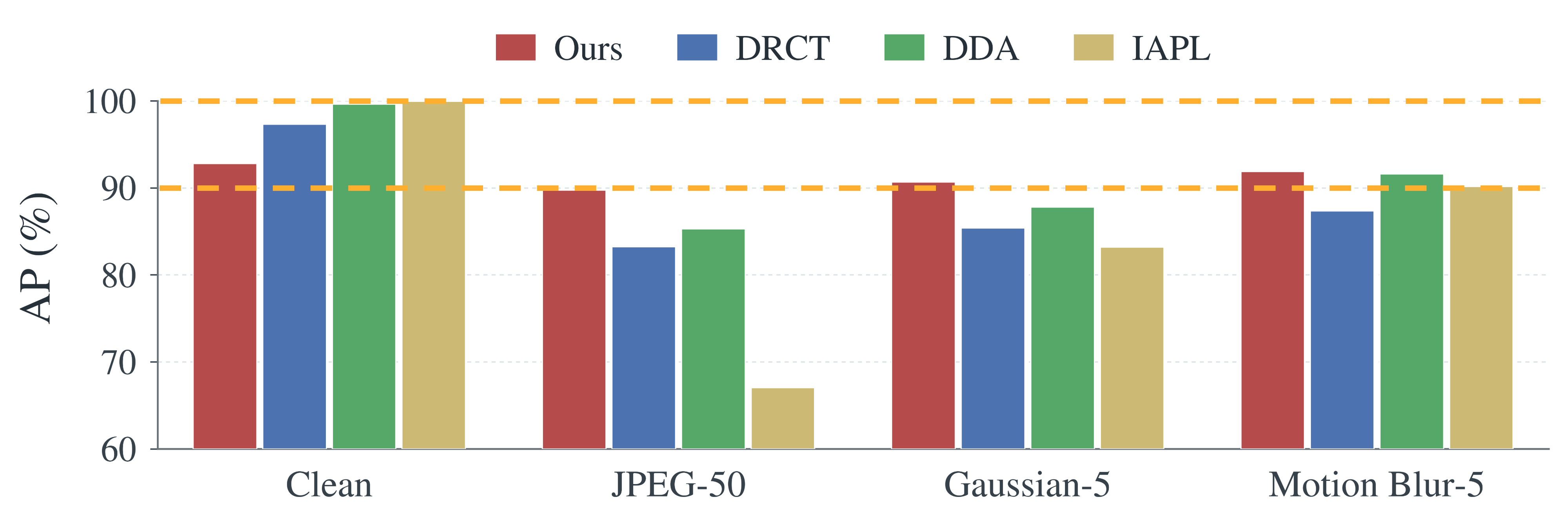}
    \caption{Robustness to perturbations on ARForensics.}
    \label{fig:race_perturbation_robustness}
\end{figure}

\begin{table}[t]
\centering
\caption{Robustness to authentic data sources on GenImage.
All metrics are reported as percentages.
$\ddagger$ denotes the authentic source used in the main experiments.
$\clubsuit$ denotes the unseen authentic source used for out-of-domain evaluation.}
\label{tab:race_real_data_robustness}
\begingroup
\setlength{\tabcolsep}{3pt}
\renewcommand{\arraystretch}{1}
\resizebox{0.75\linewidth}{!}{%
\begin{tabular}{@{}lccc@{}}
\hline\hline
Authentic data source & AUROC $\uparrow$ & AP $\uparrow$ & FPR $\downarrow$ \\
\hline
MS COCO                     & 97.10 & 96.30 & 3.96 \\
LSUN                        & 97.75 & 97.91 & 2.15 \\
ImageNet$\ddagger$          & 93.59 & 91.78 & 7.90 \\
Open Images$\clubsuit$   & 90.64 & 87.06 & 9.05 \\
\hline\hline
\end{tabular}%
}
\endgroup
\end{table}

\begin{table*}[t]
\centering
\caption{Component ablation of RACE on six datasets (mAP, \%; $\uparrow$).
Average denotes the macro-average across the six datasets.
The best result in each column is shown in \textbf{bold}.}
\label{tab:race_ablation}
\begingroup
\setlength{\tabcolsep}{5pt}
\renewcommand{\arraystretch}{1.15}
\resizebox{0.9\textwidth}{!}{%
\begin{tabular}{@{}l*{6}{c}|c@{}}
\hline\hline
Variant & GenImage & ARForensics & QwenImageBench
& EvalGEN & ForenSynths & AIGIBench & Average \\
\hline
DINO-only
& 69.97 & 58.15 & 50.85 & 53.60 & 66.29 & 57.62 & 59.41 \\
CLIP-only
& 47.87 & 44.52 & 77.03 & 52.59 & 66.72 & 76.39 & 60.85 \\
\hline
w/o PRP
& \textbf{93.13} & 91.89 & 80.79 & 86.36
& 93.25 & 93.20 & 89.77 \\
\hline
w/o Global Branch
& 91.82 & 91.30 & 80.48 & 86.57 & 92.57 & 92.89 & 89.27 \\
w/o Local Branch
& 80.28 & 76.75 & 72.07 & 88.40 & 84.81 & 77.48 & 79.97 \\
\hline
w/o RSR & 87.47 & 86.38 & 80.45 & 90.12 & 88.80 & 89.12 & 87.06 \\
\hline
\textbf{Full RACE}
& 91.78 & \textbf{92.79} & \textbf{93.45} & \textbf{98.81}
& \textbf{97.56} & \textbf{98.25} & \textbf{95.44} \\
\hline\hline
\end{tabular}%
}
\endgroup
\end{table*}

\paragraph{Datasets.}
We evaluate detection using generated images from six benchmarks: GenImage~\cite{nips/ZhuCYHLLT0H023}, ARForensics~\cite{Zhang_2025_ICCV}, QwenImageBench~\cite{Li2026QwenImageBenchFG}, EvalGEN~\cite{nips/ChenXYZWXCXGYD25}, ForenSynths~\cite{cvpr/WangW0OE20}, and AIGIBench~\cite{li2026artificial}. Authentic images are drawn from a mixed dataset comprising LSUN~\cite{yu2015lsun}, MS COCO~\cite{lin2014microsoft}, and ImageNet~\cite{deng2009imagenet}. RACE uses 5,000 authentic images for training, 1,000 for development, and an independent set of 5,000 for threshold calibration.

\paragraph{Baselines.}
We compare RACE with 13 representative detectors: CNNSpot~\cite{cvpr/WangW0OE20}, UnivFD~\cite{cvpr/OjhaLL23}, DIRE~\cite{iccv/WangBZWHCL23}, NPR~\cite{cvpr/TanLZWGLW24}, DRCT~\cite{icml/ChenZYY24}, D$^3$QE~\cite{Zhang_2025_ICCV}, LOTA~\cite{Wang_2025_ICCV}, DDA~\cite{nips/ChenXYZWXCXGYD25}, MF$^2$DA~\cite{mu2026mf2da}, IAPL~\cite{li2026iapl}, Veritas~\cite{iclr/abs-2508-21048}, RealStats~\cite{corr/abs-2601-18900}, and ConV~\cite{zhang2026detecting}. We use the publicly released weights for Veritas. The remaining supervised baselines are retrained following their official implementations, using generated images from GenImage and the mixed authentic dataset. RealStats uses the mixed authentic dataset for statistical calibration, while ConV and RACE are trained exclusively on authentic images from this dataset.

\paragraph{Implementation Details.}
RACE uses frozen CLIP~\cite{radford2021learning}, DINOv2~\cite{oquab2023dinov2}, and Noiseprint++~\cite{guillaro2023trufor} backbones. CFE and NP-RSR are trained independently for 40 epochs with AdamW~\cite{loshchilov2017decoupled}, using a learning rate of $3\times10^{-4}$ and weight decay of $10^{-4}$. PRP introduces no trainable parameters. Using 5,000 independent authentic images and a nominal false-positive budget of 5\%, we obtain a calibrated threshold of $1.896 \pm 0.007$ (mean $\pm$ standard deviation across runs). Within each run, this threshold remains fixed across all test benchmarks and generators. We report AUROC and AP, and run all experiments on NVIDIA RTX PRO 6000D GPUs.

\subsection{Performance on Image Detection Benchmarks}
\label{sec:main_results}

Table~\ref{tab:race_generalization} shows that RACE achieves the best average AUROC of 97.18\% and AP of 96.17\% across the five out-of-domain benchmarks. Although DDA and IAPL outperform RACE on their GenImage training domain, their advantage does not consistently extend to newer generators. On QwenImageBench, RACE achieves 95.34\% AUROC, compared with 81.32\% for DDA and 87.64\% for IAPL. This contrast illustrates that strong discrimination within a familiar generation domain does not necessarily provide a transferable criterion for authenticity. RACE, meanwhile, maintains AUROC between 94.69\% and 99.09\% across all five out-of-domain benchmarks without generated training examples, demonstrating that its performance extends across diverse generator distributions.

The comparison with authentic-only baselines further highlights the importance of the detection design. ConV and RealStats achieve average AUROCs of 72.45\% and 55.82\%, respectively, substantially below RACE. Their results indicate that using authentic images alone does not automatically yield strong generalization. RACE demonstrates that this setting can support competitive detection against supervised methods, suggesting that how authenticity evidence is represented and combined matters alongside the choice of training data.

Veritas provides a different comparison: it leads in both AUROC and AP on QwenImageBench and EvalGEN, but performs less well on GenImage and ARForensics, resulting in six-dataset averages of 90.51\% AUROC and 89.74\% AP. RACE's advantage lies in sustaining high performance across a broader range of benchmarks. This pattern is consistent with its joint assessment of multiple aspects of authentic-image consistency, while the following ablations examine the contributions of the individual components.

\subsection{Robustness Evaluation}
\label{sec:robustness}

\paragraph{Robustness to image perturbations.}
We evaluate robustness on ARForensics under JPEG-50, Gaussian-5, and Motion Blur-5 perturbations. As shown in Figure~\ref{fig:race_perturbation_robustness}, RACE achieves the highest AP under all three perturbations, despite trailing DRCT, DDA, and IAPL on clean images. The advantage is particularly evident under JPEG-50: RACE retains approximately 90\% AP, while IAPL drops from nearly 100\% on clean images to approximately 67\%. This ranking reversal highlights an advantage beyond clean-image performance: RACE's combined global and local evidence remains discriminative when common image degradations alter the original image statistics.

\paragraph{Robustness to authentic data sources.}
We vary the source of authentic test images while keeping the generated images from GenImage fixed. As shown in Table~\ref{tab:race_real_data_robustness}, RACE maintains AUROC above 90\% across all four sources, including the held-out Open Images~\cite{kuznetsova2020open}, where it achieves 90.64\% AUROC and 87.06\% AP. At a fixed calibrated threshold, the false-positive rate (FPR) increases from 7.90\% on ImageNet to 9.05\% on Open Images. Thus, although ranking performance declines on the held-out source, this source shift produces only a limited increase in false alarms. Transfer without source-specific recalibration suggests that the authentic-image reference captures regularities useful beyond the sources used to construct it.

\begin{figure*}[t]
    \centering
    \includegraphics[width=0.8\linewidth]{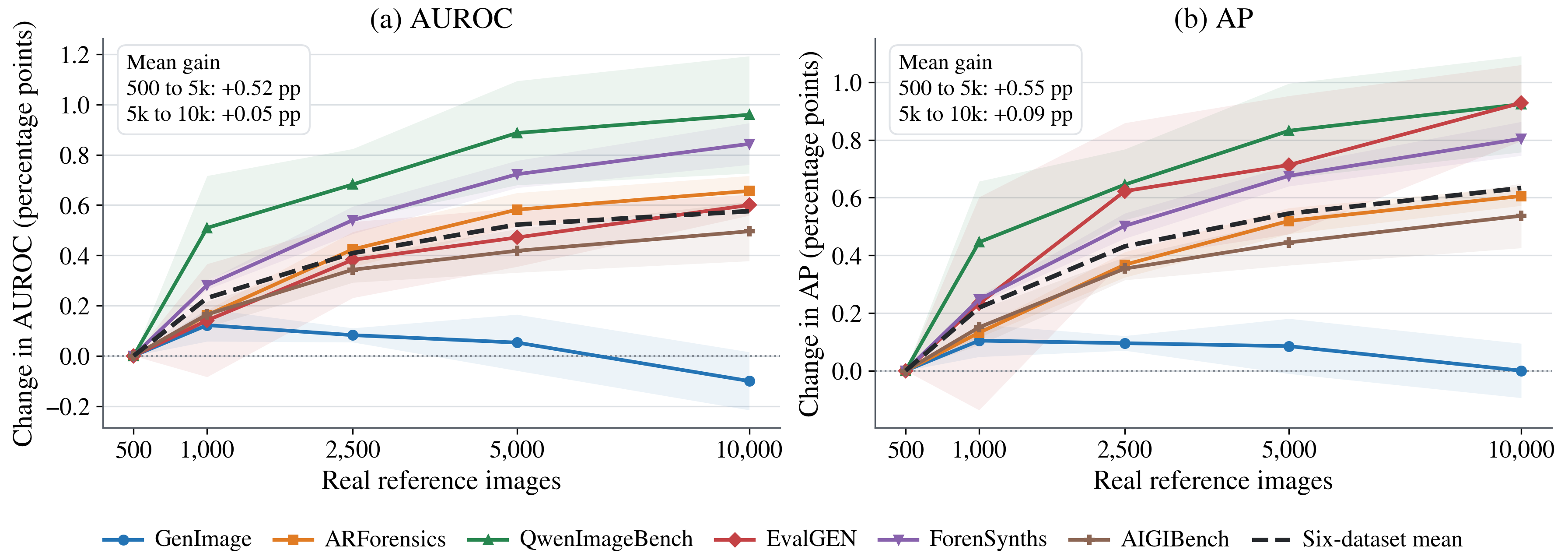}
    \caption{Sensitivity to reference-bank size.
    Changes in AUROC and AP are relative to $N=500$;
    dashed curves show the six-dataset mean.}
    \label{fig:race_reference_size}
\end{figure*}

\begin{figure*}[t]
    \centering
    \includegraphics[width=0.85\linewidth]{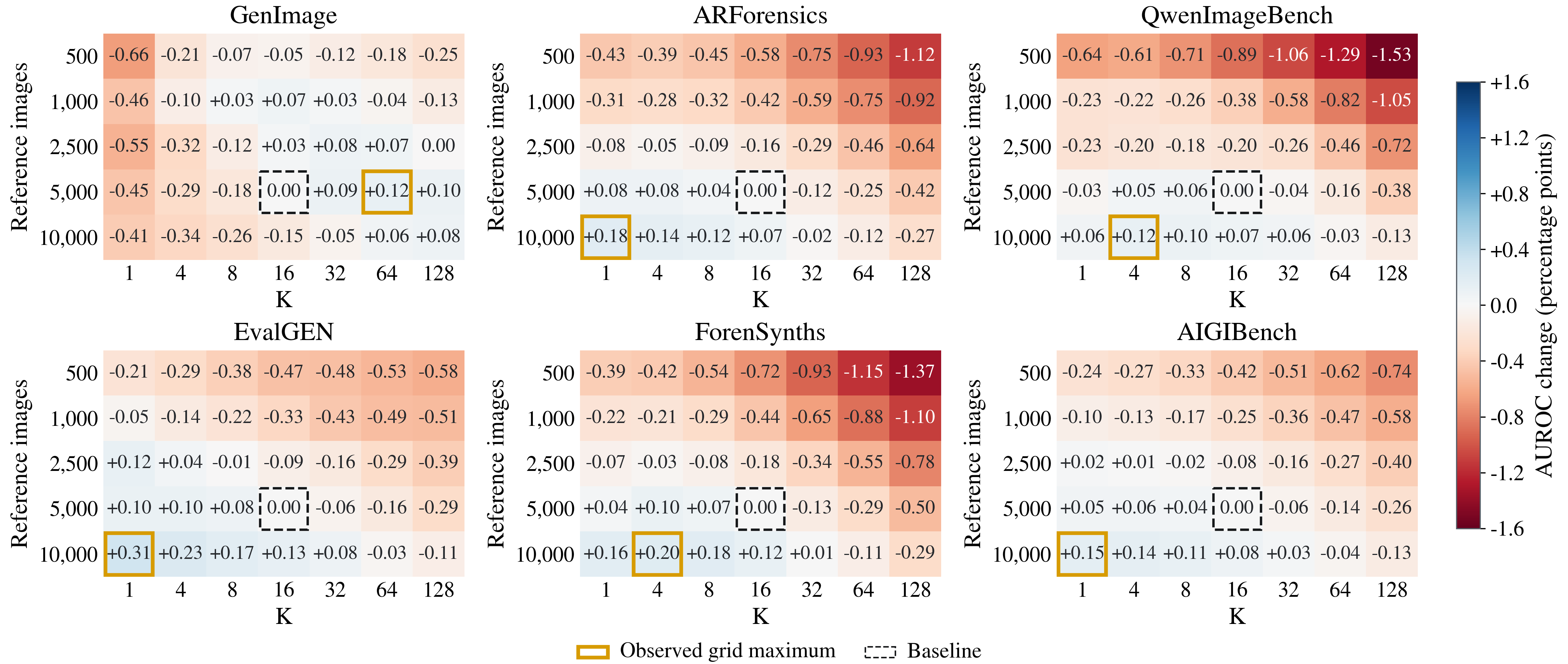}
    \caption{Joint sensitivity to reference-bank size $N$ and neighborhood size $K$.
    Values are AUROC changes in percentage points relative to $(N,K)=(5000,16)$.
    Dashed and gold boxes mark the baseline and the best observed setting in each dataset, respectively.}
    \label{fig:race_reference_k}
\end{figure*}

\subsection{Ablation Study}

\paragraph{Complementarity of encoders.}
The single-encoder variants exhibit different strengths across benchmarks: DINO-only performs better on GenImage, whereas CLIP-only leads on QwenImageBench. Full RACE outperforms both variants on all benchmarks, achieving 95.44\% average mAP compared with 59.41\% for DINO-only and 60.85\% for CLIP-only. The reversal in their relative rankings indicates that neither encoder is uniformly preferable across the evaluated distributions, motivating CFE's combination of representations with different pretrained similarity structures.

\paragraph{Contribution of PRP.}
Adding PRP raises mAP from 80.79\% to 93.45\% on QwenImageBench and from 86.36\% to 98.81\% on EvalGEN. These are the largest improvements and occur on the two weakest benchmarks of the variant without PRP. The benefit is therefore concentrated where the remaining system struggles most, suggesting that perturbation sensitivity helps recover discrimination on challenging distributions rather than providing a uniform improvement across benchmarks.

\paragraph{Asymmetric contributions of global and local branches.}
The local-only variant achieves 89.27\% average mAP, exceeding the 79.97\% of the global-only variant, while their combination reaches 95.44\%. Adding local evidence improves performance on every benchmark. In contrast, the global branch provides more selective gains, particularly on QwenImageBench and EvalGEN, with little effect on GenImage. These results reveal an asymmetric division of roles: local trace reconstruction provides the stronger standalone detector, while global assessment compensates for its weaknesses on specific datasets.

\paragraph{Effect of selective residual readout.}
The w/o RSR variant retains the same trained reconstructor but scores errors across all coordinates instead of only the 24 static terminal-residual coordinates. This change lowers mAP on all six benchmarks, reducing the average from 95.44\% to 87.06\%. Because the reconstructor is unchanged, the improvement can be attributed to the scoring rule. A complete reconstruction target therefore need not imply that aggregate reconstruction error is the most effective detection statistic.

\paragraph{Reference bank and neighborhood size.}
We fixed $N=5000$ and $K=16$ before benchmark evaluation as practical defaults. The reference-bank size was chosen to balance authentic-reference coverage against storage and retrieval costs, while a moderate neighborhood size was intended to reduce sensitivity to individual reference samples while preserving locality. We assess the sensitivity to these fixed choices in Figures~\ref{fig:race_reference_size} and~\ref{fig:race_reference_k}. Increasing $N$ beyond 5000 yields diminishing returns: expanding the bank to 10,000 doubles embedding storage for only marginal improvements in mean AUROC and AP. Although the best-performing $K$ varies across datasets, $K=16$ remains close to the best observed AUROC on all six benchmarks at $N=5000$. These results support the practicality of the fixed configuration, which is used unchanged across benchmarks.
\section{Conclusion}
We presented RACE, a framework for generalizable AI-generated image detection grounded in authentic-image regularities. Its joint global--local assessment broadens the evidence available for detection, while selective residual readout separates reconstruction context from detection evidence. Across six benchmarks, RACE achieves leading average AUROC and strong cross-generator generalization without generated training examples, alongside robustness to the evaluated image perturbations and authentic-source shifts. Our method thus balances detection performance and reference-bank storage cost.

\section{LLM Usage Statement}
OpenAI GPT models were used to assist with language polishing of the manuscript, including improvements to grammar, wording, and clarity, and to provide implementation suggestions during experimental code development. The authors reviewed and verified all AI-assisted text and code and take full responsibility for the final manuscript, implementation, and reported results.

\vfill
{
    \small
    \bibliographystyle{ieeenat_fullname}
    \bibliography{main}

@String(CVPR= {IEEE Conf. Comput. Vis. Pattern Recog.})

@String(ICCV= {Int. Conf. Comput. Vis.})

@String(ECCV= {Eur. Conf. Comput. Vis.})

@String(ICIP = {IEEE Int. Conf. Image Process.})

@String(ICLR = {Int. Conf. Learn. Represent.})

@String(CVPR  = {CVPR})

@String(ICCV  = {ICCV})

@String(ECCV  = {ECCV})

@String(ICIP  = {ICIP})

@String(ICLR  = {ICLR})

@article{corr/abs-2602-02222,
  author       = {Ruiqi Liu and
                  Manni Cui and
                  Ziheng Qin and
                  Zhiyuan Yan and
                  Ruoxin Chen and
                  Yi Han and
                  Zhiheng Li and
                  Junkai Chen and
                  ZhiJin Chen and
                  Kaiqing Lin and
                  Jialiang Shen and
                  Lubin Weng and
                  Jing Dong and
                  Yan Wang and
                  Shu Wu},
  title        = {{MIRROR:} Manifold Ideal Reference ReconstructOR for Generalizable
                  AI-Generated Image Detection},
  journal      = {CoRR},
  volume       = {abs/2602.02222},
  year         = {2026}
}

@InProceedings{corr/abs-2601-18900,
  title = 	 { RealStats: A Rigorous Real-Only Statistical Framework for Fake Image Detection },
  author =       {Zisman, Haim and Shaham, Uri},
  booktitle = 	 {AISTATS},
  pages = 	 {892--900},
  year = 	 {2026}
}

@article{corr/abs-2606-06918,
  author       = {Abhishek Ameta and
                  Sayan Banerjee and
                  Shreyas Pandith and
                  Harshit and
                  Ankita Chatterjee and
                  Akshay Janardan Bankar and
                  Amit Satish Unde},
  title        = {{DRIFT:} From Robustness Gaps to Invariance Manifolds for AI-Generated
                  Image Detection},
  journal      = {CoRR},
  volume       = {abs/2606.06918},
  year         = {2026}
}

@article{mu2026mf2da,
  author  = {Wenpeng Mu and Qiang Xu and Yining Zhang and
             Xiao Zhang and Huiyuan Wang and Yuefan Li and
             Xinghao Jiang and Tanfeng Sun},
  title   = {{MF2DA}: Multi-Level Feature Fusion for Robust Detection
             and Attribution of Universal {AI}-Generated Images},
  journal = {IEEE Trans. Inf. Forensics Security},
  year    = {2026},
}

@inproceedings{nips/ChenXYZWXCXGYD25,
  author       = {Ruoxin Chen and
                  Junwei Xi and
                  Zhiyuan Yan and
                  Ke{-}Yue Zhang and
                  Shuang Wu and
                  Jingyi Xie and
                  Xu Chen and
                  Lei Xu and
                  Isabel Guan and
                  Taiping Yao and
                  Shouhong Ding},
  title        = {Dual Data Alignment Makes AI-Generated Image Detector Easier Generalizable},
  booktitle    = {NeurIPS},
  volume       = {38},
  year         = {2025}
}

@article{series2013learning,
  title={Learning what to expect (in visual perception)},
  author={Seri{\`e}s, Peggy and Seitz, Aaron R},
  journal={Frontiers in human neuroscience},
  volume={7},
  pages={668},
  year={2013}
}

@article{rao1999predictive,
  title={Predictive coding in the visual cortex: a functional interpretation of some extra-classical receptive-field effects},
  author={Rao, Rajesh PN and Ballard, Dana H},
  journal={Nature neuroscience},
  volume={2},
  number={1},
  pages={79--87},
  year={1999}
}

@article{tenenbaum2000global,
  title={A global geometric framework for nonlinear dimensionality reduction},
  author={Tenenbaum, Joshua B and Silva, Vin de and Langford, John C},
  journal={science},
  volume={290},
  number={5500},
  pages={2319--2323},
  year={2000}
}

@inproceedings{guillaro2023trufor,
  title={Trufor: Leveraging all-round clues for trustworthy image forgery detection and localization},
  author={Guillaro, Fabrizio and Cozzolino, Davide and Sud, Avneesh and Dufour, Nicholas and Verdoliva, Luisa},
  booktitle={CVPR},
  pages={20606--20615},
  year={2023}
}

@article{10.1561/2200000101,
    author = {Angelopoulos, Anastasios N. and Bates, Stephen},
    title = {Conformal Prediction: A Gentle Introduction},
    journal = {Foundations and Trends in Machine Learning},
    volume = {16},
    number = {4},
    pages = {494-591},
    year = {2023},
}

@article{xu2025advancements,
  author  = {Xu, Qiang and Mu, Wenpeng and Li, Jianing and Sun, Tanfeng and Jiang, Xinghao},
  title   = {Advancements in AI-Generated Content Forensics: A Systematic Literature Review},
  journal = {ACM Comput. Surv.},
  volume  = {58},
  number  = {3},
  pages   = {1--36},
  year    = {2025}
}

@article{oquab2023dinov2,
  title={Dinov2: Learning robust visual features without supervision},
  author={Oquab, Maxime and Darcet, Timoth{\'e}e and Moutakanni, Th{\'e}o and Vo, Huy and Szafraniec, Marc and Khalidov, Vasil and Fernandez, Pierre and Haziza, Daniel and Massa, Francisco and El-Nouby, Alaaeldin and others},
  journal      = {CoRR},
  volume       = {abs/2304.07193},
  year={2023}
}

@inproceedings{radford2021learning,
  title={Learning transferable visual models from natural language supervision},
  author={Radford, Alec and Kim, Jong Wook and Hallacy, Chris and Ramesh, Aditya and Goh, Gabriel and Agarwal, Sandhini and Sastry, Girish and Askell, Amanda and Mishkin, Pamela and Clark, Jack and others},
  booktitle={ICML},
  pages={8748--8763},
  year={2021}
}

@article{bardes2021vicreg,
  title={Vicreg: Variance-invariance-covariance regularization for self-supervised learning},
  author={Bardes, Adrien and Ponce, Jean and LeCun, Yann},
  journal      = {CoRR},
  volume       = {abs/2105.04906},
  year={2021}
}

@inproceedings{nips/ZhuCYHLLT0H023,
  author       = {Mingjian Zhu and
                  Hanting Chen and
                  Qiangyu Yan and
                  Xudong Huang and
                  Guanyu Lin and
                  Wei Li and
                  Zhijun Tu and
                  Hailin Hu and
                  Jie Hu and
                  Yunhe Wang},
  title        = {GenImage: {A} Million-Scale Benchmark for Detecting AI-Generated Image},
  booktitle    = {NeurIPS},
  volume       = {36},
  year         = {2023},
}

@InProceedings{Zhang_2025_ICCV,
    author    = {Zhang, Yanran and Yu, Bingyao and Zheng, Yu and Zheng, Wenzhao and Duan, Yueqi and Chen, Lei and Zhou, Jie and Lu, Jiwen},
    title     = {D3QE: Learning Discrete Distribution Discrepancy-aware Quantization Error for Autoregressive-Generated Image Detection},
    booktitle = {ICCV},
    year      = {2025},
    pages     = {16292--16301}
}

@article{Li2026QwenImageBenchFG,
  author        = {Niantong Li and Guangzheng Hu and Weixu Qiao and
                   Ying Ba and Qichen Hong and Shijun Shen and
                   Jinlin Wang and Fan Zhou and Jianye Kang and
                   Xin Shang and Ziyi He and Wei Wang and Dalin Li and
                   Jiahao Li and Jie Zhang and Kaiyuan Gao and Kun Yan and
                   Lihan Jiang and Ningyuan Tang and Shengming Yin and
                   Tianhe Wu and Xiao Xu and Xiaoyue Chen and
                   Yuxiang Chen and Yan Shu and Yanran Zhang and
                   Yilei Chen and Yixian Xu and Zekai Zhang and
                   Zhendong Wang and Zihao Liu and Zikai Zhou and
                   Hongzhu Shi and Yi Wang and Bing Zhao and Hu Wei and
                   Lin Qu and Chenfei Wu},
  title         = {{Qwen-Image-Bench}: From Generation to Creation
                   in Text-to-Image Evaluation},
  journal      = {CoRR},
  volume       = {abs/2605.28091},
  year          = {2026}
}

@inproceedings{cvpr/WangW0OE20,
  author       = {Sheng{-}Yu Wang and
                  Oliver Wang and
                  Richard Zhang and
                  Andrew Owens and
                  Alexei A. Efros},
  title        = {CNN-Generated Images Are Surprisingly Easy to Spot... for Now},
  booktitle    = {CVPR},
  pages        = {8692--8701},
  year         = {2020},
}

@article{li2026artificial,
  title={Is artificial intelligence generated image detection a solved problem?},
  author={Li, Ziqiang and Yan, Jiazhen and He, Ziwen and Zeng, Kai and Jiang, Weiwei and Xiong, Lizhi and Fu, Zhangjie},
  journal={NeurIPS},
  volume={38},
  year={2026}
}

@article{yu2015lsun,
  title={Lsun: Construction of a large-scale image dataset using deep learning with humans in the loop},
  author={Yu, Fisher and Seff, Ari and Zhang, Yinda and Song, Shuran and Funkhouser, Thomas and Xiao, Jianxiong},
  journal      = {CoRR},
  volume       = {abs/1506.03365},
  year={2015}
}

@inproceedings{lin2014microsoft,
  title={Microsoft coco: Common objects in context},
  author={Lin, Tsung-Yi and Maire, Michael and Belongie, Serge and Hays, James and Perona, Pietro and Ramanan, Deva and Doll{\'a}r, Piotr and Zitnick, C Lawrence},
  booktitle={ECCV},
  pages={740--755},
  year={2014}
}

@inproceedings{deng2009imagenet,
  title={Imagenet: A large-scale hierarchical image database},
  author={Deng, Jia and Dong, Wei and Socher, Richard and Li, Li-Jia and Li, Kai and Fei-Fei, Li},
  booktitle={CVPR},
  pages={248--255},
  year={2009}
}

@inproceedings{cvpr/OjhaLL23,
  author       = {Utkarsh Ojha and
                  Yuheng Li and
                  Yong Jae Lee},
  title        = {Towards Universal Fake Image Detectors that Generalize Across Generative
                  Models},
  booktitle    = {CVPR},
  pages        = {24480--24489},
  year         = {2023},
}

@inproceedings{iccv/WangBZWHCL23,
  author       = {Zhendong Wang and
                  Jianmin Bao and
                  Wengang Zhou and
                  Weilun Wang and
                  Hezhen Hu and
                  Hong Chen and
                  Houqiang Li},
  title        = {{DIRE} for Diffusion-Generated Image Detection},
  booktitle    = {ICCV},
  pages        = {22388--22398},
  year         = {2023},
}

@inproceedings{cvpr/TanLZWGLW24,
  author       = {Chuangchuang Tan and
                  Huan Liu and
                  Yao Zhao and
                  Shikui Wei and
                  Guanghua Gu and
                  Ping Liu and
                  Yunchao Wei},
  title        = {Rethinking the Up-Sampling Operations in CNN-Based Generative Network
                  for Generalizable Deepfake Detection},
  booktitle    = {CVPR},
  pages        = {28130--28139},
  year         = {2024},
}

@inproceedings{icml/ChenZYY24,
  author       = {Baoying Chen and
                  Jishen Zeng and
                  Jianquan Yang and
                  Rui Yang},
  title        = {{DRCT:} Diffusion Reconstruction Contrastive Training towards Universal
                  Detection of Diffusion Generated Images},
  booktitle    = {ICML},
  pages        = {7621--7639},
  year         = {2024}
}

@InProceedings{Wang_2025_ICCV,
    author    = {Wang, Hongsong and Cheng, Renxi and Zhang, Yang and Han, Chaolei and Gui, Jie},
    title     = {LOTA: Bit-Planes Guided AI-Generated Image Detection},
    booktitle = {ICCV},
    year      = {2025},
    pages     = {17246--17255}
}

@inproceedings{li2026iapl,
  author    = {Yiheng Li and Zichang Tan and Guoqing Xu and
               Zhen Lei and Xu Zhou and Yang Yang},
  title     = {Towards Generalizable {AI}-Generated Image Detection
               via Image-Adaptive Prompt Learning},
  booktitle = {CVPR},
  year      = {2026},
  pages     = {21262--21272}
}

@inproceedings{iclr/abs-2508-21048,
  author       = {Hao Tan and
                  Jun Lan and
                  Zichang Tan and
                  Ajian Liu and
                  Chuanbiao Song and
                  Senyuan Shi and
                  Huijia Zhu and
                  Weiqiang Wang and
                  Jun Wan and
                  Zhen Lei},
  title        = {Veritas: Generalizable Deepfake Detection via Pattern-Aware Reasoning},
  booktitle    = {ICLR},
  year         = {2026}
}

@article{zhang2026detecting,
  title={Detecting generated images by fitting natural image distributions},
  author={Zhang, Yonggang and Nie, Jun and Tian, Xinmei and Gong, Mingming and Zhang, Kun and Han, Bo},
  journal={NeurIPS},
  volume={38},
  pages={31337--31361},
  year={2026}
}

@article{loshchilov2017decoupled,
  title={Decoupled weight decay regularization},
  author={Loshchilov, Ilya and Hutter, Frank},
  journal      = {CoRR},
  volume       = {abs/1711.05101},
  year={2017}
}

@article{kuznetsova2020open,
  title={The open images dataset v4: Unified image classification, object detection, and visual relationship detection at scale},
  author={Kuznetsova, Alina and Rom, Hassan and Alldrin, Neil and Uijlings, Jasper and Krasin, Ivan and Pont-Tuset, Jordi and Kamali, Shahab and Popov, Stefan and Malloci, Matteo and Kolesnikov, Alexander and others},
  journal={International journal of computer vision},
  volume={128},
  number={7},
  pages={1956--1981},
  year={2020}
}

@inproceedings{iclr/PopeZAGG21,
  author       = {Phillip Pope and
                  Chen Zhu and
                  Ahmed Abdelkader and
                  Micah Goldblum and
                  Tom Goldstein},
  title        = {The Intrinsic Dimension of Images and Its Impact on Learning},
  booktitle    = {ICLR},
  year         = {2021},
}

@inproceedings{iclr/YanLCHJ0X25,
  author       = {Shilin Yan and
                  Ouxiang Li and
                  Jiayin Cai and
                  Yanbin Hao and
                  Xiaolong Jiang and
                  Yao Hu and
                  Weidi Xie},
  title        = {A Sanity Check for AI-generated Image Detection},
  booktitle    = {ICLR},
  year         = {2025},
}

@inproceedings{icip/McCloskeyA19,
  author       = {Scott McCloskey and
                  Michael Albright},
  title        = {Detecting GAN-Generated Imagery Using Saturation Cues},
  booktitle    = {ICIP},
  pages        = {4584--4588},
  year         = {2019},
}

@inproceedings{NatarajMMCFBR19,
  author       = {Lakshmanan Nataraj and
                  Tajuddin Manhar Mohammed and
                  B. S. Manjunath and
                  Shivkumar Chandrasekaran and
                  Arjuna Flenner and
                  Jawadul H. Bappy and
                  Amit K. Roy{-}Chowdhury},
  title        = {Detecting {GAN} generated Fake Images using Co-occurrence Matrices},
  booktitle    = {Media Watermarking, Security, and Forensics},
  year         = {2019},
}

@article{corr/abs-2405-20112,
  author       = {Zhiyuan He and
                  Pin{-}Yu Chen and
                  Tsung{-}Yi Ho},
  title        = {{RIGID:} {A} Training-free and Model-Agnostic Framework for Robust
                  AI-Generated Image Detection},
  journal      = {CoRR},
  volume       = {abs/2405.20112},
  year         = {2024}
}
}


\end{document}